\documentclass[letterpaper]{article} 
\usepackage{aaai24}  
\usepackage{times}  
\usepackage{helvet}  
\usepackage{courier}  
\usepackage[hyphens]{url}  
\usepackage{graphicx} 
\usepackage{natbib}  
\usepackage{caption} 
\usepackage[ruled,linesnumbered,noend]{algorithm2e}
\usepackage{eqparbox}
\usepackage{stackengine}
\newcommand{\mike}[1]{
\textcolor{blue}{}
}
\newcommand{\updated}[1]{#1}

\usepackage{float}
\usepackage{listings}
\DeclareCaptionStyle{ruled}{labelfont=normalfont,labelsep=colon,strut=off} 
\floatstyle{ruled}
\newfloat{listing}{tb}{lst}{}
\floatname{listing}{Listing}
\title{Traffic Flow Optimisation for Lifelong Multi-Agent Path Finding }
\author {Zhe Chen\textsuperscript{\rm 1}, Daniel Harabor\textsuperscript{\rm 1}, Jiaoyang Li\textsuperscript{\rm 2}, Peter J. Stuckey\textsuperscript{\rm 1,3}
}
\affiliations {
\textsuperscript{\rm 1}Department of Data Science and Artificial Intelligence, Monash University, Melbourne, Australia\\
\textsuperscript{\rm 2}Robotics Institute, Carnegie Mellon University, Pittsburgh, USA \\
\textsuperscript{\rm 3}OPTIMA Australian Research Council ITTC, Melbourne, Australia\\
\{zhe.chen,daniel.harabor,peter.stuckey\}@monash.edu, jiaoyangli@cmu.edu
}

\usepackage{bibentry}

\usepackage{adjustbox}
\usepackage{appendix}
\usepackage{amsmath}
\usepackage{amssymb}
\usepackage{amsfonts}
\usepackage{amsthm}
\usepackage{thmtools}
\usepackage{subcaption,dcolumn}
\usepackage{tikz}
\usetikzlibrary{arrows,decorations,fit,positioning,shapes}
\usepackage{empheq,fancybox}
\usepackage{adjustbox}
\usepackage{appendix}
\usepackage{stmaryrd}
\usepackage{multirow}
\usepackage{bm}
\usepackage{xcolor}         
\usepackage{booktabs}       
\usepackage{url}            
\usepackage[english]{babel}
\newcommand*\rot[1]{\rotatebox{90}{#1}}

\newtheorem{example}{Example}

\newtheorem*{summary*}{Summary}

\declaretheoremstyle[
headpunct={},
headfont=\bfseries,
notefont=\bfseries,
bodyfont=\normalfont,
headformat=\NAME~\NUMBER:\NOTE.
]{def}

\newcommand{\ignore}[1]{}

\def\squareforqed{\hbox{\rlap{$\sqcap$}$\sqcup$}}
\def\qed{\ifmmode\squareforqed\else{\unskip\nobreak\hfil
\penalty50\hskip1em\null\nobreak\hfil\squareforqed
\parfillskip=0pt\finalhyphendemerits=0\endgraf}\fi}

\newcommand{\true}{\mathit{true}}
\newcommand{\false}{\mathit{false}}

\definecolor{tyellow1}{HTML}{FCE94F}
\definecolor{tyellow2}{HTML}{EDD400}
\definecolor{tyellow3}{HTML}{C4A000}
\definecolor{torange1}{HTML}{FCAF3E}
\definecolor{torange2}{HTML}{F57900}
\definecolor{torange3}{HTML}{C35C00}
\definecolor{tbrown1}{HTML}{E9B96E}
\definecolor{tbrown2}{HTML}{C17D11}
\definecolor{tbrown3}{HTML}{8F5902}
\definecolor{tgreen1}{HTML}{8AE234}
\definecolor{tgreen2}{HTML}{73D216}
\definecolor{tgreen3}{HTML}{4E9A06}
\definecolor{tblue1}{HTML}{729FCF}
\definecolor{tblue2}{HTML}{3465A4}
\definecolor{tblue3}{HTML}{204A87}
\definecolor{tpurple1}{HTML}{AD7FA8}
\definecolor{tpurple2}{HTML}{75507B}
\definecolor{tpurple3}{HTML}{5C3566}
\definecolor{tred1}{HTML}{EF2929}
\definecolor{tred2}{HTML}{CC0000}
\definecolor{tred3}{HTML}{A40000}
\definecolor{tlgray1}{HTML}{EEEEEC}
\definecolor{tlgray2}{HTML}{D3D7CF}
\definecolor{tlgray3}{HTML}{BABDB6}
\definecolor{tdgray1}{HTML}{888A85}
\definecolor{tdgray2}{HTML}{555753}
\definecolor{tdgray3}{HTML}{2E3436}

\addto\extrasenglish{%

}

\graphicspath{{plots/},{plots/28,28_mnist_1v3_flip},{plots/28,28_mnist_1v7_flip},{plots/10,10_mnist_1v3_flip},{plots/10,10_mnist_1v7_flip}}

\DeclareGraphicsExtensions{.pdf}

\newcommand{\old}{\phi}
\newcommand{\new}{\theta}

\begin{document}

\maketitle
\begin{abstract}
Multi-Agent Path Finding (MAPF) is a fundamental problem in robotics that asks us to compute collision-free paths for a team of agents, all moving across a shared map. Although many works appear on this topic, all current algorithms struggle as the number of agents grows.  The principal reason is that existing approaches typically plan free-flow optimal paths, which creates congestion. To tackle this issue, we propose a new approach for MAPF where agents are guided to their destination by following congestion-avoiding paths.
We evaluate the idea in two large-scale settings: one-shot MAPF, where each agent has a single destination, and lifelong MAPF, where agents are continuously assigned new destinations. Empirically, we report large improvements in solution quality for one-short MAPF and in overall throughput for lifelong MAPF.

\end{abstract}
\section{Introduction} \label{sec:intro}

Multi-Agent Path Finding (MAPF)~\cite{stern2019multi} is a fundamental coordination problem in robotics and also 
at the heart of many industrial applications. For example, in automated fulfilment~\cite{wurman2008coordinating} and sortation centres~\cite{kou2020idle}, 
a team of robots must work together to deliver packages. To complete such tasks, robots must navigate across a shared map to reach their goal locations.  Robots must arrive at their goals as quickly as possible, and they must do so while 
avoiding collisions with static obstacles in the environment as well as with other moving robots. 
In the classical, sometimes called {\em one-shot}, MAPF problem~\cite{stern2019multi}, robots are modelled as simplified agents, 
with each being assigned a single goal location.  
A related setup, known as {\em lifelong} MAPF~\cite{MaAAMAS17,li2021lifelong}, continuously assigns new goal locations to 
agents as they arrive at their current goal locations. 

Both one-shot and lifelong MAPF problems have been intensely studied, with a variety
of substantial advancements reported in the literature. For example, leading optimal 
~\cite{sharon2015conflict,LiAIJ21,LiIJCAI19,LamCOR22,GangeICAPS19} and bounded-suboptimal~\cite{li2021eecbs} MAPF
algorithms now scale to hundreds of agents, while providing solution quality guarantees. 
Yet some real applications require up to {\em thousands of simultaneous agents}, and at this scale, only unbounded suboptimal approaches are currently applicable.

Two leading frameworks for unbounded suboptimal MAPF are Large Neighbourhood Search (LNS)~\cite{LiIJCAI21,LiAAAI22,li2021scalable} and Priority Inheritance with Back Tracking (PIBT)~\cite{okumura2022priority,okumura2023improving}. 
Approaches based on LNS modify an existing (potentially infeasible) MAPF plan by iteratively changing the 
paths for a few agents to reduce the number of collisions or their travel times. A main drawback of LNS-based methods is that the performance of the single-agent path planner significantly degrades as map size and number of  
agents increase.  This restricts the applicability of LNS as, when paired with an aggressive timeout (often required for real applications),  timeout failure can occur before any iterations are completed. 
Meanwhile, PIBT-based approaches
use rule-based collision avoidance to plan paths. They compute paths timestep by timestep, 
which is extremely efficient. 
For this reason, PIBT-based methods usually scale to substantially more agents than LNS-based methods. 
However, this type of planners guides agents toward their goals using individually optimal \emph{free-flow} heuristics (i.e., considering only travel distance while ignoring other agents), a strategy known to create high levels of congestion. 
Moreover, its computed solutions tend to have higher costs than those computed by LNS-based methods~\cite{shen2023tracking}. 

A related body of work on the Traffic Assignment Problem
(TAP)~\cite{TAP} 
from the transportation community considers similar issues. 
Like MAPF, TAP approaches compute optimised paths for agents moving across a shared map. 
In TAP, multiple agents can use the same edge at the same time, but travel time increases with the number of agents. TAP algorithms thus try to compute a User Equilibrium (UE) solution, where no agent can improve their arrival time by switching to a different path with lower cost.
Inspired by such ideas, we propose to compute time-independent routes for MAPF agents, which take into account {\em expected 
congestion} due to other agents following similar trajectories. 
We use the resulting \emph{guide paths} as improved heuristics for PIBT, which allows fast planning for large numbers of agents.
Yet we also continuously improve the guide paths, like LNS, which serves to further increase the quality of computed plans. 
Experimental results show convincing improvements for one-shot and lifelong MAPF against a range of
leading algorithms, including PIBT, LNS and LaCAM.

\section{Problem Definition} \label{sec:prelim}

The Multi-Agent Path Finding (MAPF) problem takes as \textbf{input} is an undirected gridmap $G = (V,E)$ and a set of $k$ agents $\{a_1...a_k\}$, where 
each agend begins at a start location $s_i \in V$ and must reach a goal location $g_i \in V$. 
Agents can move from one grid cell (equiv. vertex) to another adjacent grid cell using one of four possible \textbf{grid moves}:
{\em North, South, East} and {\em West}.
Time is discretised into unit-sized steps.  At each timestep $t$, every agent must move to an adjacent grid cell or else {\em wait} in place. 
Each action (wait or move) has a cost of 1.
We use the notation $(v_i, t_j)$ to say that vertex $v$ is occupied by agent $i$ at timestep $t_j$.

A \textbf{vertex conflict}, $\langle a_i$, $a_j$, $v$, $t\rangle$, occurs when two agents $a_i$ and $a_j$ occupy the same vertex $v \in V$ at the same timestep $t$. 
An {\bf edge conflict}, $\langle a_i$, $a_j$, $e$, $t\rangle$, occurs when two agents $a_i$ and $a_j$ pass through the same edge $e \in E$ in opposite directions at 
the same timestep $t$. 
A \textbf{path} is a sequence of actions that transitions an agent from a vertex $v_i$ to another vertex $v_j$. 
We say that the path is valid (equiv. feasible or collision-free) if no action produces an edge or vertex conflict.
The \textbf{cost} of a path is the sum of its action costs. 

A \textbf{solution} to a MAPF problem is a collision-free path assignment that allows every agent $i$ to move from $s_i$ to $g_i$
and then remain at $g_i$ without any conflict. 
The \textbf{objective} in MAPF is to find a feasible solution that minimises the \emph{Sum of Individual (path) Costs} (SIC).

{\bf Lifelong Multi-Agent Path Finding:}
in this variant, agents are assigned sequences of tasks that must be completed. A task for agent $i$ is a request
to visit a goal location $g_i$. Unlike MAPF, when $g_i$ is reached, the agent is not required to wait indefinitely. Instead, 
agent $i$ is assigned a new goal $g'_i$. The problem continues indefinitely with 
the objective being to maximise total task completions by an operational time limit {\em T} (i.e., 
max {\em throughput}).
Note that we assume the \emph{task assigner} is external to our path-planning system and that agents only know their current goal location.


\section{Background}
We briefly summarise previous works which are necessary for understanding our main contributions.
\subsection{FOCAL Search}
This 
is a best-first bounded-suboptimal algorithm~\cite{pearl1982studies} similar to $A^*$. It uses an \emph{OPEN} list to prioritise node expansions in the usual $f=g+h$ ordering. 
A second priority queue, \emph{FOCAL}, contains only those nodes from {\em OPEN} with $f \leq w \cdot f_{min}$, where $f_{min}$ is the minimal $f$-value of any node in {\em OPEN} and $w \geq 1$ is a user-specified acceptance criterion that indicates the maximum suboptimality of an admissible solution. 
Since every node in \emph{FOCAL} is admissible, they can be sorted with any alternative criteria, denoted $\hat{h}$.
This approach guarantees solutions of cost not more than $w \cdot C^*$, where $C^*$ is the optimal cost.
Notice that $A^*$ is a special case of FOCAL Search where $w=1$.

\IncMargin{0.2em}
\begin{algorithm}[t]
    \SetKwProg{Fn}{}{}{end}
    $p' \leftarrow p$; // Create a copy of initial priority $p$\;
    \Fn{\textbf{PlanStep}($A,\old,p,p'$)}{
        \updated{\For{$a_i \in A$}{ 
            $\new[a_i] \leftarrow \bot $; //$\bot$ means no action decided\;
            $g_i = \old[a_i]$ $?$  $p_i \leftarrow p'_i$ : $p_i \leftarrow p_i + 1$; \label{alg:pibt:lowest_priority}
            }
        sort $A$ in decreasing order of $p_i \in p$ \;}
        \For{$a_i \in A$}{\label{alg:pibt:begin}
            \If{$\new[a_i] = \bot$} {
                \textbf{PIBT}($a_i,\bot, \old, \new, A$);\label{alg:pibt:base_step}
            }  
        } 
        \Return $\new$ \label{alg:pibt:end}\;
    }
    ~\\
    \Fn{\textbf{PIBT}($a_i, a_j, \old, \new, A$)}{
        $C \leftarrow \{ v ~|~ (\old[a_i], v) \in E \}$\;\label{alg:pibt:sort_begin}
        sort $C$ based on increasing $dist(v,g_i)$\;\label{alg:pibt:sort_end}
        \For{$v \in C$ in order}{
        \lIf{$\exists a \in A, \new[a] = v$} {continue} \label{alg:pibt:vertex_collision}
        \lIf{$a_j \neq \bot \wedge \old[a_j] = v$}{continue} \label{alg:pibt:edge_collision}
        $\new[a_i] \leftarrow v$\; 
        \If{$\exists a \in A, \old[a] = v \wedge \new[a] = \bot$} {
            \lIf{$\neg$ \textsf{PIBT}($a,a_i,\old,\new,A)$}{continue} \label{alg:pibt:push}
        }
            \Return $\true$\;
        }
        $\new[a_i] \leftarrow \old[a_i]$\;
        \Return $\false$\;
    }
    \caption{PIBT. In each iteration {\em PlanStep} computes a next move $\new$ for each agent $a \in A$, currently at position $\old$, using a priority ordering $p$.} 
    \label{alg:PIBT}
\end{algorithm}

\subsection{PIBT and LaCAM} 
PIBT
is an iterative rule-based approach for solving MAPF problems~\cite{okumura2022priority}. 
The basic schema, which we adopt in this work, is sketched in Algorithm~\ref{alg:PIBT}. 
\updated{In each iteration or timestep (lines~\ref{alg:pibt:begin}-\ref{alg:pibt:end}) all agents
plan a single step $\new$ toward their goal locations by calling the PIBT function.
Moves are selected according to the individual shortest distance to agents' goal locations, where the move toward a location with shorter distances to the goal location is preferred, 
with the next agent to plan being selected according to an initial priority order $p$.}
When two agents compete for the same next location, a simple 1-step reservation scheme is applied. It allows higher-priority agents 
to reserve their next moves while lower-priority agents are bumped to other less desirable locations. 
\updated{This strategy is applied recursively, which means that bumped agents are selected next, thus inheriting the priority of the higher priority agent.}(line~\ref{alg:pibt:push}).
Once every agent is planned the moves are executed. In other words, agents advance toward their goal locations and a new iteration begins. 

PIBT guarantees that the highest priority agent will eventually reach its goal location, at which point it becomes the lowest priority agent 
(line~\ref{alg:pibt:lowest_priority}).
Thus PIBT never produces deadlock situations. However, in one-shot MAPF, the PIBT strategy may still produce livelocks,
which can occur among pairs of agents where the goal position of one appears on the planned path of the other.  This makes PIBT incomplete in general.  
In lifelong MAPF the situation is different. Here arriving agents are immediately assigned new tasks, which resolves any contention. 
Thus for lifelong MAPF the PIBT algorithm is indeed complete.

{\bf LaCAM*}: is an anytime search strategy that combines a systematic joint-space search with PIBT to compute suboptimal MAPF plans more efficiently~\cite{okumura2023improving}.  Each node of LaCAM* is a configuration of the agents on the map.  Successor nodes are generated by invoking PIBT, with the move selected for each agent being restricted by a
set of associated constraints. LaCAM systematically explores the set of joint-space plans but uses a type of partial expansion known as Operator Decomposition~\cite{standley2010finding} to avoid the usual branching factor explosion. 
The addition of a systematic search allows LaCAM* to succeed more often than PIBT, and to compute higher-quality plans,
while retaining its performance advantages.

\section{Traffic Congestion Reasoning for MAPF}

\begin{figure}[t!]
\centering
\includegraphics[width=0.6\columnwidth]{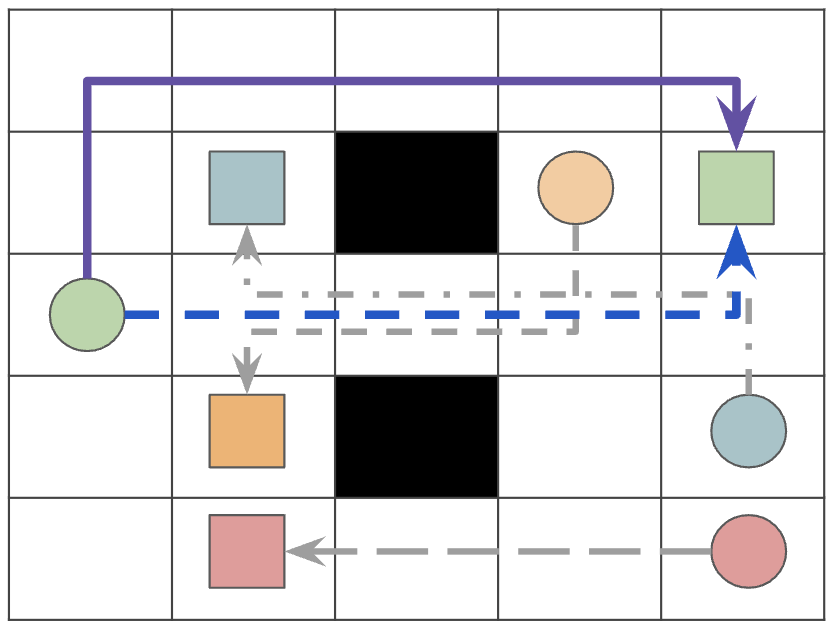}
\caption{\label{fig:path_comparison}
We show a small MAPF problem with 4 agents. Dashed (blue) lines indicate individually optimal paths from each $s_i$ to each $g_i$. Solid (purple) lines indicate congestion-aware individually-optimal paths.}
\end{figure}

A simple approach to planning each agent in a MAPF problem is to simply generate the shortest path for each agent independently of all the rest. This is the policy employed by PIBT and it has two main advantages: (i) the paths are \emph{time independent}, 
which makes them fast to compute; (ii) the path costs provide strong {\em lower-bounds} on the true time-dependent optimal cost, which makes these paths useful as guiding heuristics, for undertaking further state-space search, such as LaCAM*. The main drawback is that following 
these paths almost always leads agents into conflicts, as the vertices that appear on time-independent paths tend to have a greater \emph{betweenness centrality}~\cite{freeman1977set} than other vertices in the graph. In other words, a vertex appearing on one time-independent optimal path is more likely to also appear on a great many other optimal paths. 

To tackle this issue we propose to compute time-independent optimal paths which are {\em congestion-aware}. 
In particular, after computing one time-independent shortest path, we update the edge costs of the graph, so that 
the edges which appear on that path become more expensive to traverse. A similar update strategy drives 
solution methods of Traffic Assignment Problems, where edge-cost increases reflect additional  {\em traffic cost}  from multiple vehicles using the same road link at the same time. 
We adapt these general ideas for MAPF by first defining a model for computing traffic costs and then
applying these costs to compute new congestion-aware move policies and {\em guide paths} for PIBT. 
Figure~\ref{fig:path_comparison} shows a representative example; we compare the set of individually optimal paths for a 
small group of agents and their corresponding congestion-aware counterparts.

%

\subsection{Traffic Costs}

Define $f_{v_1,v_2}$ as the \emph{flow} from vertex $v_1$ to $v_2$ given a set of agents
whose time-independent shortest paths each have the state $v_1$ followed by $v_2$. 
\updated{Thus $f_{v_1,v_2}$ is the number of agents traverse the edge $(v_1,v_2)$ from $v_1$ to $v_2$ direction in the current path assignment. Note that $f_{v_2,v_1}$ indicates traversing through the same edge but from $v_2$ to $v_1$ direction.
} 

Define the \emph{vertex congestion} $c_v$ of vertex $v \in V$ as 
$c_v = \frac{n \times (n-1)}{2}$ where $n = \sum_{v' \in V : (v',v) \in E} f_{v',v}$
is the total number of agents entering vertex $v$.
$c_v$ represents the least total delay that will occur assuming all agents enter the vertex at the same time, since
in the best case each agent will have to wait for all the agents preceding it. Apportioning this congestion to each agent equally leads to a cost per agent of $p_v = \lceil \frac{c_v}{n} \rceil = \lceil \frac{n-1}{2} \rceil$.

\begin{figure}[t!]
\centering
\includegraphics[width=0.7\columnwidth]{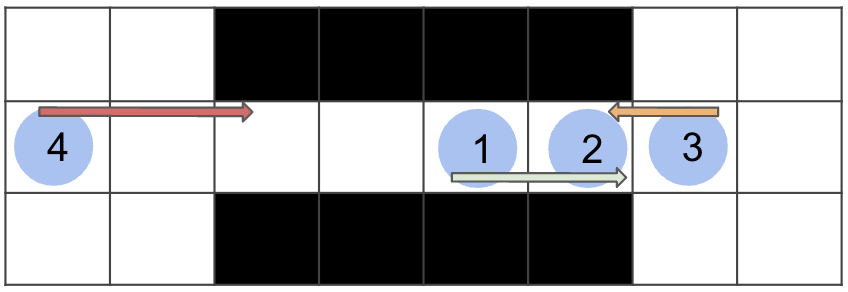}
\caption{\label{fig:contraflow} In this example high-priority agents enter a
corridor just before a set of lower-priority agents exit, \updated{the number on each agent indicating its priority.} 
Each time this occurs there is a large increase in the objective function value. }
\end{figure}

Define the \emph{contraflow congestion} $c_e$ of undirected edge $e \equiv (v_1, v_2) \in E$ as $c_e = f_{v_1,v_2} \times f_{v_2,v_1}$. This formulation reflects our observation that PIBT can push agents of higher priority into a corridor which then forces lower-priority agents, already inside the corridor, to reverse direction. 
The result is an exponential increase in path cost for all agents attempting to cross. Figure \ref{fig:contraflow} shows an example.

Traffic Assignment Problem uses congestion charges to directly modify affected edge weights and then replan shortest paths. This is also beneficial for MAPF but we found the {\em actual cost} of contraflow congestion is massively underestimated by our model above. To overcome this issue we generate two-part edge weights to more accurately model congestion. \updated{Each edge $e = (v_1,v_2)$ is given a two-part weighted cost $(c_e, 1 + p_{v_2})$}, where 1 indicates the free-flow (i.e., zero congestion) cost of using edge $e$. We then search for (lexicographically) shortest paths for each agent using this two-part cost, i.e. first minimising contraflow costs, and then weighted edge costs. 
\updated{
 In addition, we explore other variants, such as $1 + c_e + p_{v_2}$ and $1 + p_{v_2}$, and comparable methods from existing literature~\cite{han2022optimizing} to compute edge weights in the experimental section.
}


\subsection{Path Planning and Improvement} 

Analogous to the process for solving the TAP~\cite{chen2002faster}, \updated{we start with \emph{flow} of 0 for each edge on each traversing direction}, and compute shortest paths iteratively. After computing one shortest path we immediately update affected edge costs. This means
the next agent uses all available traffic information when computing its shortest path.
The algorithm is formalised in Algorithm~\ref{alg:tpaths}. 

\begin{algorithm}[t]
    \SetKwProg{Fn}{}{}{end}
    \SetKw{Call}{}

    \Fn{\textbf{FindPaths}($A,V,E,s,g$)}{
        \updated{\For{$(v_1,v_2) \in E$}{
        $f[v_1,v_2] \leftarrow 0$;$f[v_2,v_1] \leftarrow 0$\;
        }} \label{alg:tpaths:init}
        \For{$a \in A$}{
            $\pi[a] \leftarrow$ \Call{\textsf{SP}($s_a,g_a,f,V,E,w$)}\;
            \For{$i \in 2..|\pi[a]|$}{
                $f[\pi[a][i-1],\pi[a][i]] ++$\;\label{alg:tpaths:counts}
            }
        }
    \Return $\pi$ \\    
    }
    ~\\
    \Fn{\textbf{PathRefinement($\pi$)}}{\label{alg:tpath:refine_begin}
        \While{not meet termination condition}{
            Choose subset $S \subset A$ \;
            $\pi \leftarrow$ \Call{\textsf{Replan}($\pi,S,A,f,V,E)$} \; 
        }\label{alg:tpath:refine_end}
    }
        
    ~\\
    \Fn{\textbf{Replan}($\pi,S, A, f,V,E$)}{\label{alg:tpath:replan_begin}
        \For{$a \in S$}{
        \For{$i \in 2..|\pi[a]|$}{
        $f[\pi[a][i-1],\pi[a][i]] --$ \;
        }
        }
        \For{$a \in S$}{
        $\pi[a] \leftarrow$ \Call{\textsf{SP}($s_a,g_a,f,V,E, w$)}\;
        \For{$i \in 2..|\pi[a]|$}{$f[\pi[a][i-1],\pi[a][i]] ++$\;}
        }
        \Return $\pi$        
    }\label{alg:tpath:replan_end}
\caption{\label{alg:tpaths} We compute (approximate) user equilibrium paths $\pi$ for a set of MAPF agents $A$ on map $(V,E)$.
The set of start locations is denoted $s$, and goal locations $g$. 
We use FOCAL Search to compute paths, with parameter $w \geq 1$ indicating the admissibility criteria for suboptimal
solutions.}
\end{algorithm}

We begin by initialising flow counts to zero (line~\ref{alg:tpaths:init}), and then plan agents one by one. We use the current flow counts $f[\cdot,\cdot]$ to compute the two-part edge weight $(c_e,1+p_v)$ for edge $e \equiv (u,v)$ during the shortest path computation $\textsf{SP}$. 
The subsequent path for agent $a$, denoted $\pi[a]$, is a sequence of vertices starting from $\pi[a][1] = s_a$ where $|\pi[a]|$ is its length (number of actions) and $\pi[a][|\pi[a]|] = g_a$.
We then update flow counts (line~\ref{alg:tpaths:counts}) and plan the next agent.

Once each agent has a path we call the procedure \textsf{PathRefinement} (lines~\ref{alg:tpath:refine_begin}-\ref{alg:tpath:refine_end}). 
This iterated method selects (at random) a subset of agents $|S|$ whose paths will be replanned, so as to improve their
goal arrival time.  The function \textsf{Replan} (lines~\ref{alg:tpath:replan_begin}-\ref{alg:tpath:replan_end}) removes 
the paths of selected agents from the stored flows and computes a new path for each agent $a \in S$ 
followed by updating flows, continuing until a termination condition is met (= max iterations).


We end with a \emph{time independent} path for each agent, that tries to take into account likely delays caused by congestion. 
Next, we discuss how these paths can be used to improve PIBT performance for one-shot and lifelong MAPF.

\subsection{Guide Paths and Guide Heuristics}

\begin{figure}[t!]
\centering
\includegraphics[width=0.65\columnwidth]{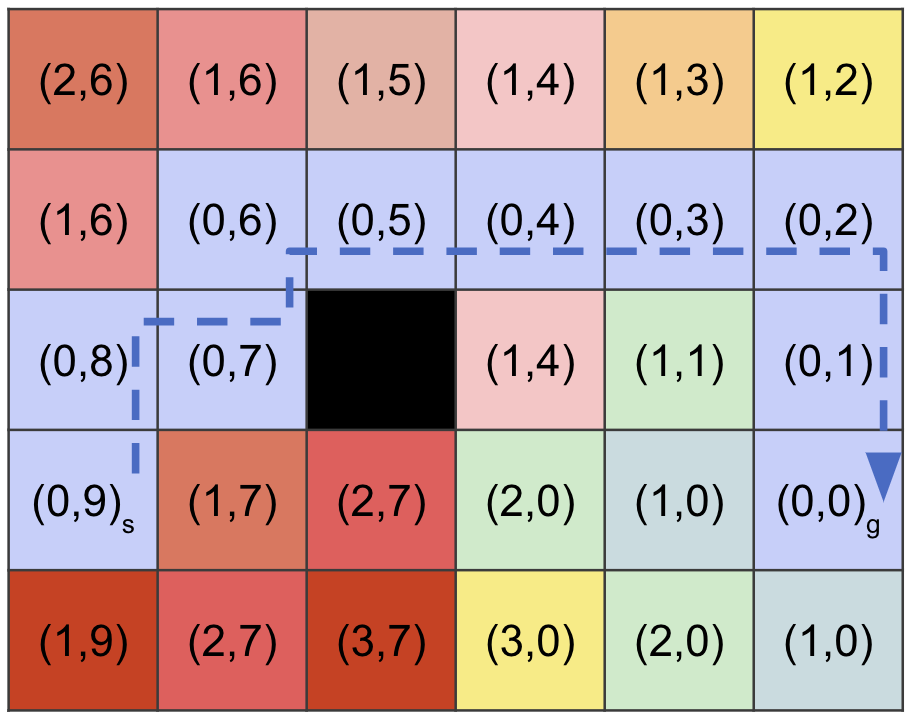}
\caption{\label{fig:guide_path}
A small figure with a guide path (the blue dashed line) and marked heuristic values.}
\end{figure}

In PIBT the free-flow distance (i.e., not considering traffic costs) from each vertex to the goal is used to decide the preferred movement direction for each agent. The agent will always choose this move unless prevented by a higher priority agent, in which case it selects the next-best move.
Since this policy does not consider the paths of any other agent, this approach invariably leads to high congestion.

In this work, we propose to modify PIBT so that each agent tries to follow a congestion-aware \emph{guide path}. For each agent $a_i$ we thus compute a \emph{guide heuristic} $h_i(v)$. Given a vertex $v \in V$ we compute a two-part value $h_i(v) = (dp, dg)$, where $dp$ is the shortest free-flow distance, from $v$ to the guide path. 
The value $dg$ meanwhile is the shortest remaining distance to the goal $g_i$, as reached subsequently by strictly following 
the guide path.
To efficiently compute guide heuristics we use a lazy backward Breadth First Search (BFS). Every location on the input guide path is pushed to the BFS queue as a root-level node. We then expand nodes lazily, layer by layer, up to the current requested location $v$. 
We cache the $h_i$-value of each expanded node along the way. When we receive a request for a non-cached $h_i$-value 
the search can be resumed. The heuristic is updated (re-computed) only when the agent receives a new goal location.
Notice that the computation of the guide heuristic does not consider traffic costs. This is because they are already reflected by the guide path, which is given as input. Meanwhile, the computed $h_i$ values are optimal because every action has cost=1.
Applying the guide heuristic in PIBT is trivial: we simply sort the adjacent vertices $C$ by $h_i(v)$ rather than $dist(v, g_i)$.

\begin{example}
Figure~\ref{fig:guide_path} shows guided planning with PIBT. Agent $a_i$ is currently at vertex $v$. Its move preference is toward an adjacent vertex $v'$ where $h_i(v')$ is lexicographically least. 
When the agent is not on the guide path its preferred move is the one that minimises the number of 
steps to reach a vertex on the guide path.  In the case of ties, the agent prefers to move toward a vertex on the guide 
path with the minimum remaining distance to the goal. 
\end{example}

In one recent and related work~\cite{han2022optimizing} authors also compute a per-vertex traffic cost metric, 
for use as a distance/heuristic tiebreaker in prioritised planning with $A^*$ search. In other words, when planning the path of each agent the search tie-breaks on vertices with lower traffic cost, provided those vertices also minimise the shortest distance to the goal. The main drawback of this approach is that the objective function does not explicitly consider traffic costs, which diminishes its effectiveness. 

\begin{algorithm}[t]
    \SetKwProg{Fn}{}{}{end}
    \SetKw{Call}{}
    
    \Fn{\textbf{GuidedPlanStep}(A,$\pi$, t, g, $\old$)}
    {
        \CommentSty{//Initialising}\;\label{alg:lifelong:init_begin}
        
        \If{$\exists a \in A, \pi[a]=\emptyset$}{
            A'$\leftarrow$A\;
            \lIf{Relax}
            { A'$\leftarrow$ \textsf{SelectNotInitialized}(A)}
            $\pi \leftarrow$ \Call{\textsf{FindPaths}(A',V,E,$\old$,g)}\;
        }\label{alg:lifelong:init_end}
        \CommentSty{//Updating}\;

        \label{alg:lifelong:update_begin}
        \For{$\forall a \in A, g[a]~changed, \pi[a] \neq \emptyset$}
        {
             $\pi \leftarrow$ \Call{\textsf{Replan}($\pi$, {a}, A,f,V,E)}\;
        }
        \label{alg:lifelong:update_end}

        \CommentSty{//Refining}\; \label{alg:lifelong:refine_begin}
        
        \lIf{$\forall a\in A, \pi[a] \neq \emptyset~\wedge$  Refine}{\Call{\textsf{PathRefinement}($\pi$)}}
        \lFor{$\forall a \in A$ where $\pi[a]$ has changed}
        {
             $h_a \leftarrow$ \textsf{Get($\pi[a]$)}
        }\label{alg:lifelong:refine_end}
        
        \CommentSty{//PIBT with \emph{Guide Heuristic}}\;
        
        $\new \leftarrow$ \Call{\textsf{PlanStep(A)}}\;\label{alg:lifelong:pibt}
        
        \Return $\new$\;
    }

\caption{\label{alg:lifelong} Lifelong Procedure to determine the next move $\new$ for each agent $a \in A$ from current position $\old$, with current goal vitices $g$, a guide heuristic table $h$, and a set of guide path $\pi$.}

\end{algorithm}

\subsection{Lifelong Procedure}

In lifelong MAPF the arrival of an agent at its goal makes redundant its guiding heuristic. 
The arrival of other agents meanwhile causes guidance data to become outdated, as traffic costs change.
To handle these situations we propose to compute guidance data continuously and entirely online. 
Algorithm~\ref{alg:lifelong} shows the corresponding procedure.

At the start of a lifelong problem, we need to compute guide paths and guide heuristics for all the agents.
Undertaken sequentially, these operations can require up to dozens of seconds,  during which time agents are 
typically assumed not to move. To smooth the response time over the operation period we propose a lazy initialisation 
scheme (lines~\ref{alg:lifelong:init_begin}-\ref{alg:lifelong:init_end}), where guidance is only computed for a 
certain number of agents per timestep. Agents without guide paths simply follow their individual shortest distance (i.e., unmodified PIBT) until they can be processed at a later timestep.

Whenever any agent is assigned a new task guidance data for all agents becomes outdated to various degrees. 
For this reason, we call the \textsf{PathRefine} procedure (Algorithm~\ref{alg:tpaths}) every timestep.  
If guidance paths for any agent have changed as a result, we compute new heuristics for those agents (lines~\ref{alg:lifelong:refine_begin}-\ref{alg:lifelong:refine_end}).
The refining of the guide paths can be undertaken even if no new tasks are assigned to any agent.
One advantage of this strategy is that guide paths can be recomputed based on agents' current locations, 
which allows us to remove flow costs due to past trajectories.
After refinement we plan agents using PIBT (line~\ref{alg:lifelong:pibt}, with the distance function replaced by 
guide heuristics if available.

\section{Experimental Results} \label{sec:res}

\begin{figure*}[t!]
\centering
\begin{subfigure}{1\textwidth}
\centering
  \includegraphics[width=0.8\columnwidth]{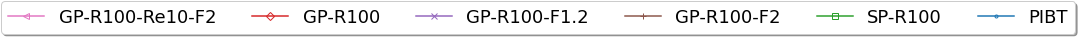}
\end{subfigure}
\begin{subfigure}{0.255\textwidth}
\centering
  \stackinset{r}{0\textwidth}{t}{-.03\textwidth}
  {\includegraphics[width=0.25\columnwidth]{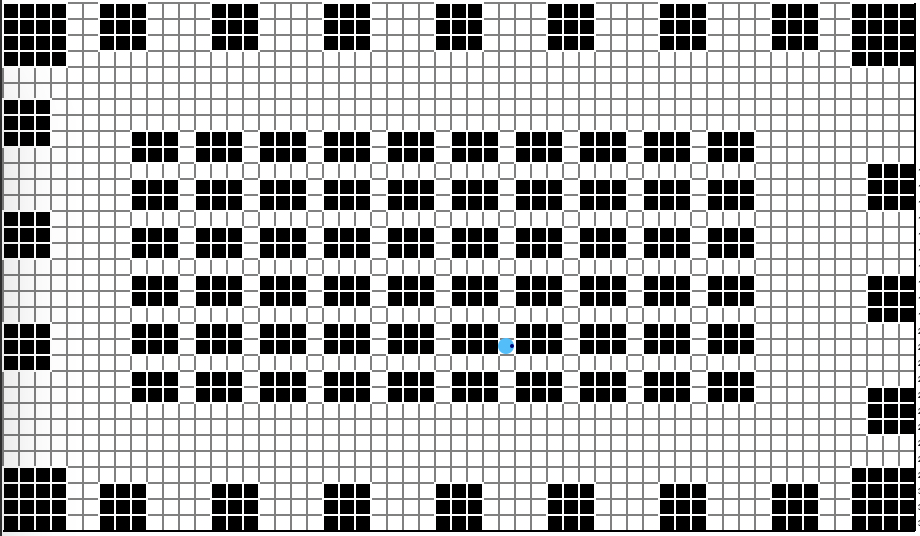}}
  {\includegraphics[width=1\columnwidth]{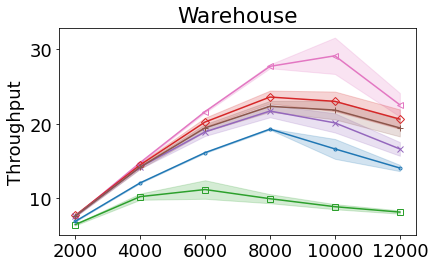}}
  \includegraphics[width=1\columnwidth]{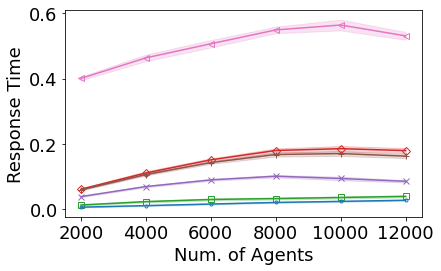}
\end{subfigure}
\begin{subfigure}{0.24\textwidth}
\centering
\stackinset{r}{0\textwidth}{t}{-.03\textwidth}
  {\includegraphics[width=0.3\columnwidth]{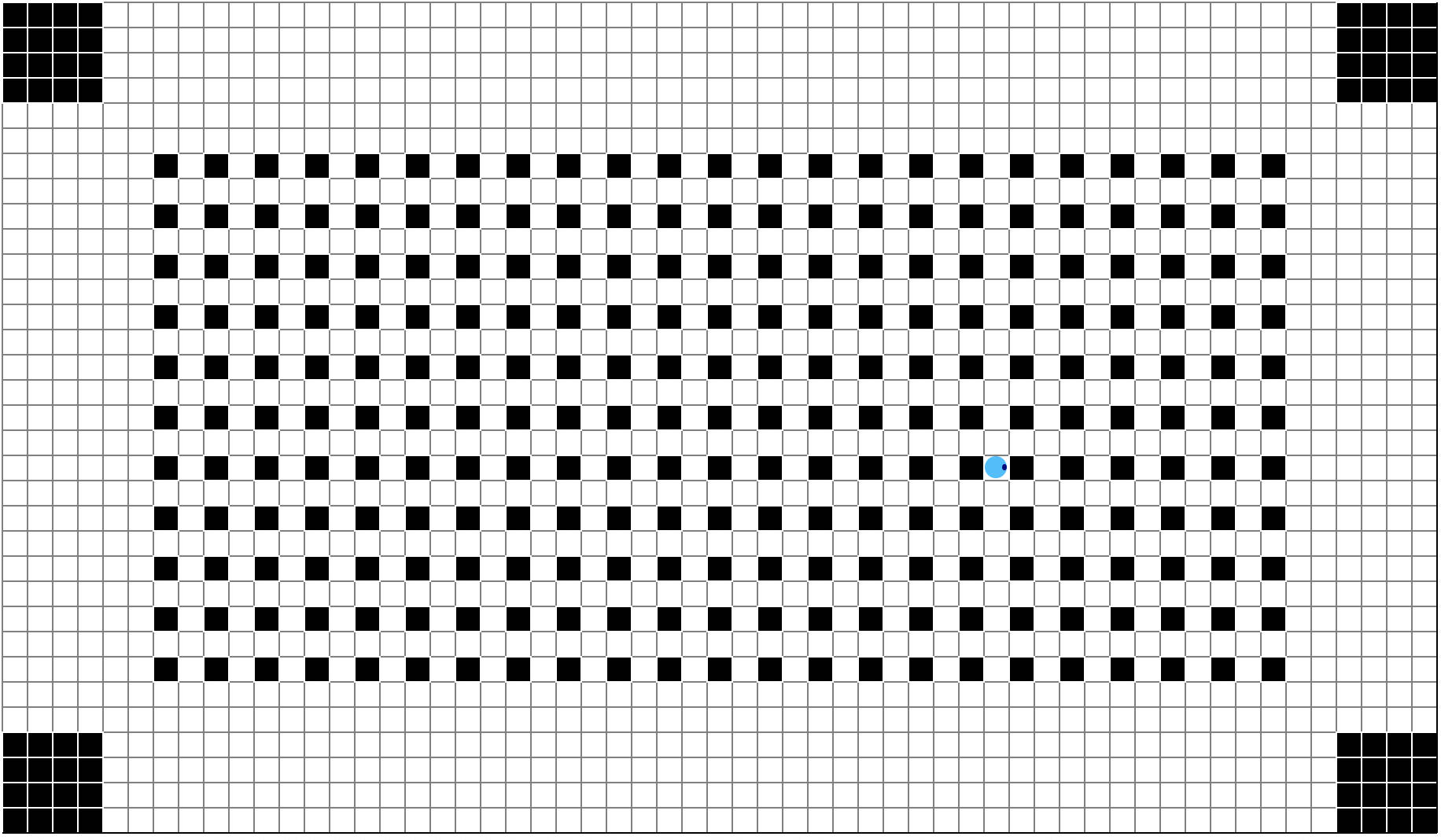}}
  {\includegraphics[width=1\columnwidth]{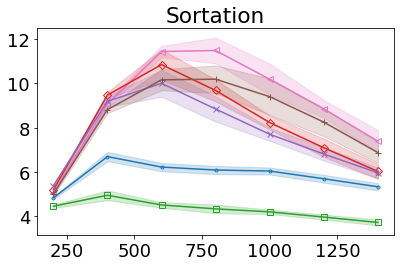}}
  \includegraphics[width=1\columnwidth]{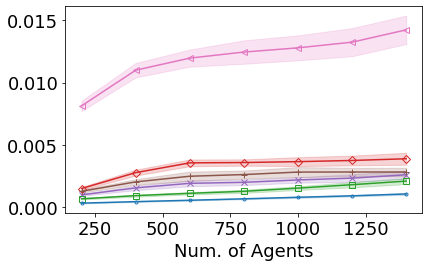}
\end{subfigure}
\begin{subfigure}{0.24\textwidth}
\centering
\stackinset{r}{.05\textwidth}{t}{-.03\textwidth}
  {\includegraphics[width=0.2\columnwidth]{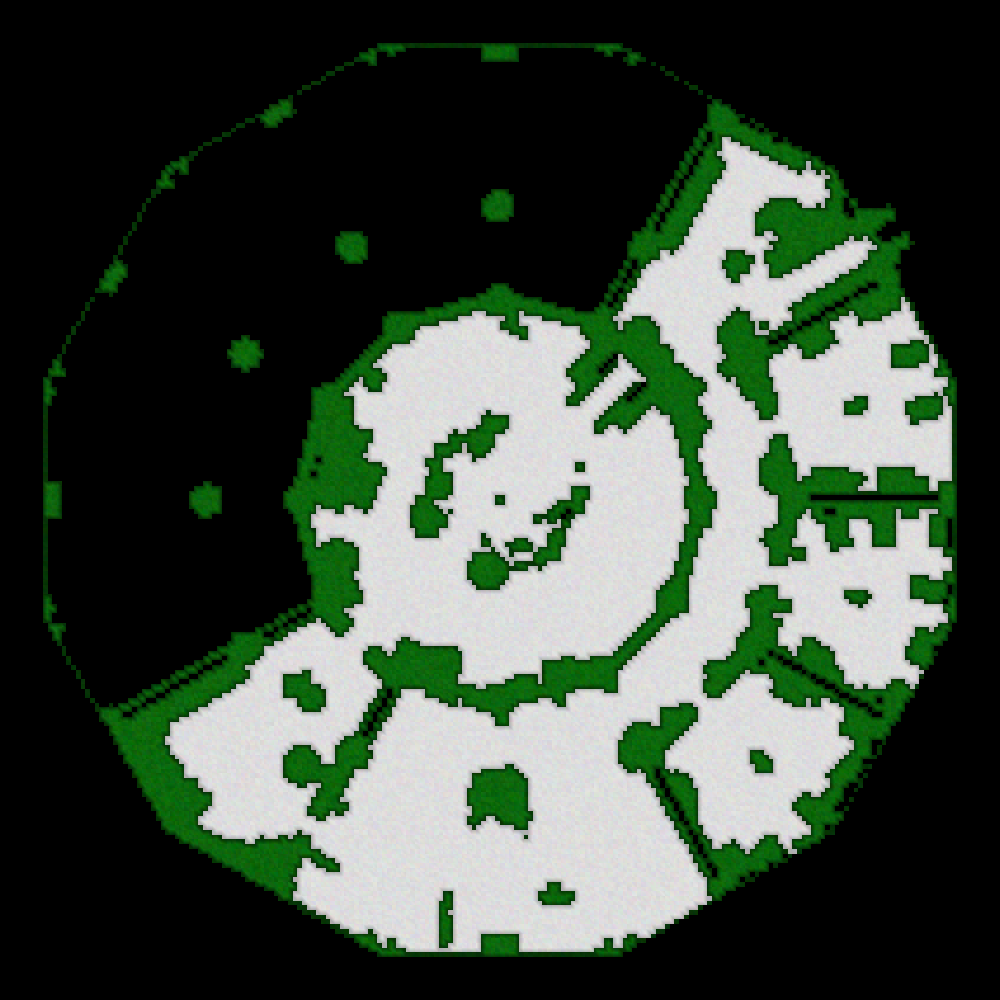}}
  {\includegraphics[width=1\columnwidth]{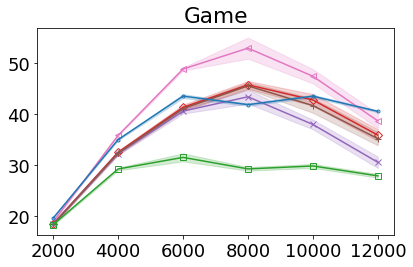}}
  \includegraphics[width=1\columnwidth]{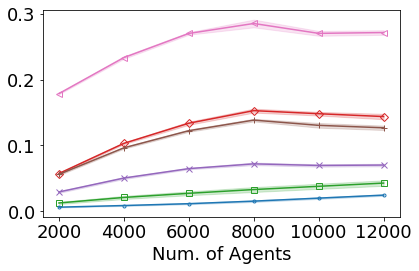}
\end{subfigure}
\begin{subfigure}{0.245\textwidth}
\centering
\stackinset{r}{.05\textwidth}{t}{-.03\textwidth}     {\includegraphics[width=0.2\columnwidth]{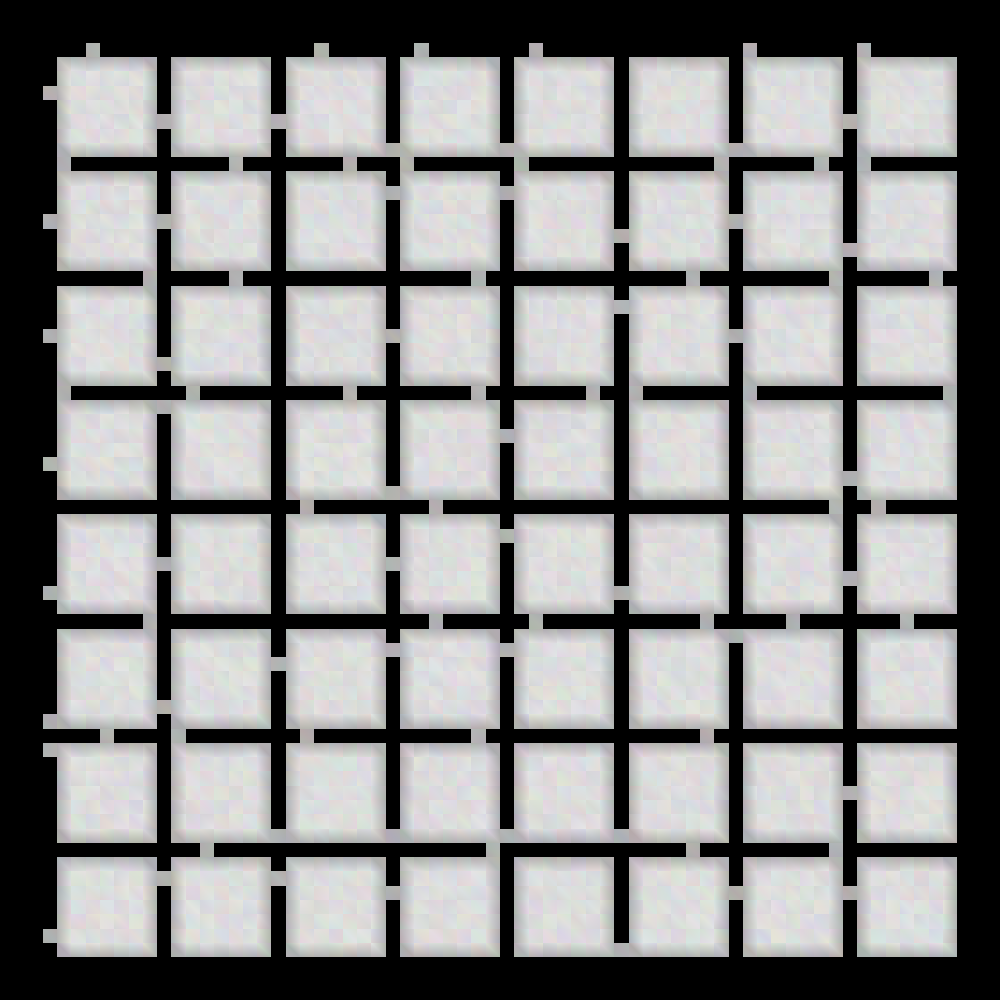}}
  {\includegraphics[width=1\columnwidth]{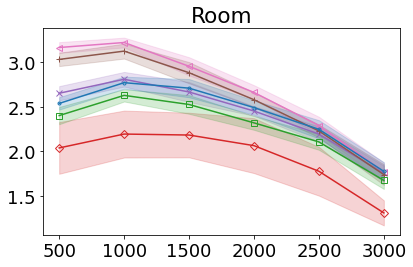}}
  \includegraphics[width=1\columnwidth]{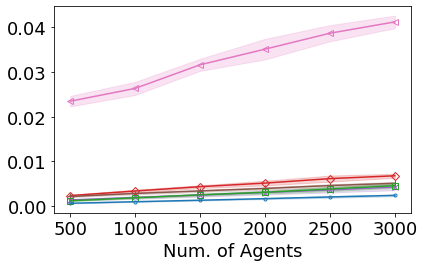}
\end{subfigure}
\caption{Lifelong MAPF. Average throughput (top) and response time in second (bottom). Shaded regions show standard deviation.} 
\label{fig:allmap}
\end{figure*}

\begin{figure*}[t!]
\centering
\includegraphics[width=0.5\linewidth]{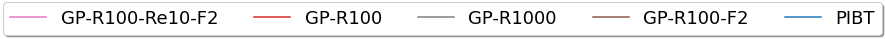}
\includegraphics[width=1\linewidth]{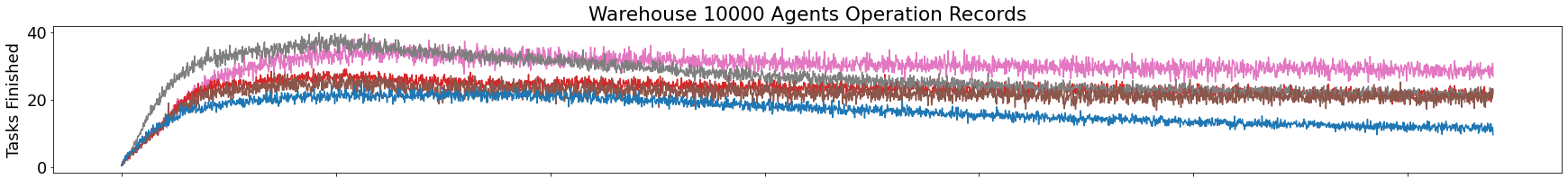}
\includegraphics[width=1\linewidth]{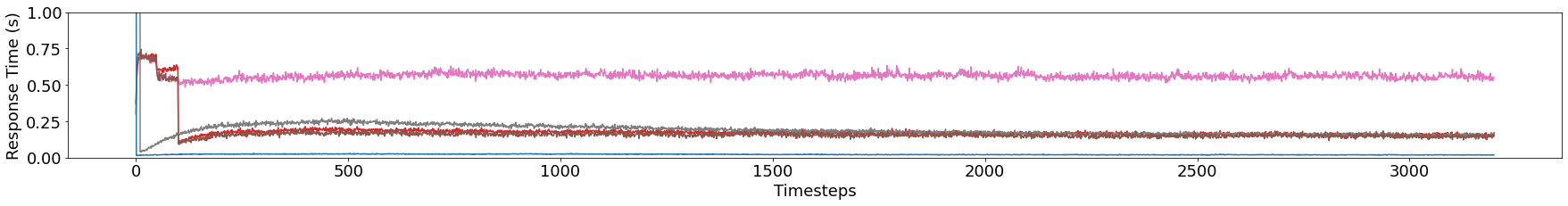}

\caption{Lifelong MAPF. Average tasks finished (top) and average response time (in the range 0-1 second; bottom). 24 instances, with 10,000 agents each. PIBT and GP-R1000 incur setup costs of 6.26 and 6.98 seconds to compute heuristic data.}
\label{fig:-record}
\end{figure*}

In our main experiment, we compare guided PIBT against the original baseline for Lifelong MAPF. We also integrate guidance 
into (one-shot) LaCAM* and compare against the original baseline for one-shot MAPF. 
Implementations\footnote{https://github.com/nobodyczcz/Guided-PIBT}  are written in C++ and evaluated on a Nectar Cloud VM instance with 32 AMD EPYC-Rome CPUs and 64 GB RAM. 

We run large-scale experiments with up to 12 thousand agents across 4 distinct map types. For each map and each number of agents, we evaluate 24 randomly sampled instances, resulting in 600 problem instances in total. 
For Lifelong MAPF, the maximum simulation time is based on the size of the map: we compute a maximum number of timesteps as 
$(with + height) \times 5$ with the intention that each agent has the opportunity to complete approximately 5 tasks. Our maps are:
\begin{itemize}
    \item \emph{Warehouse}: a $500 \times 140$ synthetic fulfillment center map with 38589 traversable cells. There are 144 instances, with between 2000 to 12000 agents (which occupy 31\% traversable cells). Simulation time: 3200 timesteps. 
    \item \emph{Sortation}: a $33 \times 57$ synthetic sortation centre map with 1564 traversable cells. There are 168 instances, with between 200 to 1400 agents (which occupy 92\% traversable cells). Simulation time: 450 timesteps. 
    \item \emph{Game}: ost003d, a $194 \times 194$ map with 13214 traversable cells, from video games. There are 144 instances with between 2000 to 12000 agents (which occupy 90\% traversable cells). Simulation time: 1940 timesteps. 
    \item \emph{Room}: room-64-64-8, a $64 \times 64$ synthetic map with 3232 traversable cells. There are 144 instances, with between 500 to 3000 agents (which occupy 93\% traversable cells). Simulation time: 640 timesteps.
\end{itemize}
In one-shot MAPF experiments, planners have 60 seconds timelimit. In lifelong MAPF experiments, planners have 10 seconds to return actions for all agents at every timestep. Failing to return in any timestep leads to timeout failure.

\subsection{Results for Lifelong MAPF}

The basic implementation of our algorithm is \texttt{GP-R100},
where GP indicates guide path heuristic guidance and \texttt{R$x$} indicates the initialization of guide paths $x$ is limited to $x$ paths per timestep. 
Algorithm names with \texttt{F$w$} indicate the algorithm uses \emph{Focal Search} to bound the path length up to 
$w \cdot C^*$. Names with \texttt{Re{$i$}} means $i$ iterations of online refinement are applied at each timestep. 
For each online refinement iteration, 10 agents are selected by two selection methods, one randomly selects agents from all agents, and the other one selects the agent with the highest traffic cost on the guide path and other agents with guide paths intersecting with it. 
We use adaptive LNS \cite{LiIJCAI21} to record the two methods' relative success in improving the solutions and choosing the most promising method to select agents for replan.
\texttt{SP} is a reference algorithm using individual shortest paths to compute guide heuristics, ignoring traffic cost.

{\bf PIBT with Guide Path:}
Figure~\ref{fig:allmap} measures the average throughput and average response time, which is the average time the planner returns actions for all agents at each timestep. 
The base implementation \texttt{GP-R100} shows a great advantage over PIBT on throughput with Warehouse and Sortation maps on all numbers of agents.
Performance is lower for Room as 
this map has many single-length corridors. It is challenging to form efficient two-way traffic here as fewer alternative paths with similar costs exist.
With \emph{Focal Search} enabled, \texttt{GP-R100-F2} exploits the additional slack to improve throughput on Room, whereas \texttt{GP-R100-F1.2} does not help as the restriction is too tight.

By enabling the online refinement, \texttt{GP-R100-Re10-F2} refines 10 iterations on 100 paths every timestep. 
This bumps the throughput on every map on every number of agents. 
We noticed that in lifelong MAPF there is a peak agent density, beyond which adding more agents decreases throughput due to increasingly severe congestion. Compared with PIBT, our methods shifted this peak to the right on each map, with \texttt{GP-R100-Re10-F2} improving Warehouse and Game by 2000 agents and Sortation by 200 agents.


{\bf Steady Operation: }
Figure \ref{fig:-record} shows timestep breakdown records for 10,000 agents on Warehouse over 24 instances. 
Initialisation for PIBT requires 6 seconds at the start of the operation while  
\texttt{GP-R1000} requires 7 seconds each for the first 10 timesteps, with 1000 guide paths being 
initialised per timestep. 
Clearly, guide paths are more expensive to compute than PIBT unit-cost distance tables (7ms per agent versus 0.6 ms per agent on the Warehouse map). This highlights the importance of lazy initialisation.

\texttt{GP-R100} and \texttt{GP-R1100-Re10-F2} have steady response times, within 1 second, for the entire planning episode. 
Interestingly, PIBT and \texttt{GP-R1000} start with an advantage for initialisation their tasks finished and soon hit a peak and then drop for the remainder of the episode, as more agents are pushed to congested areas. 
\texttt{GP-R100-Re10-F2} remains steady throughout. 

\begin{table*}[t]
\centering
\begin{tabular}{l|l|l|l|l|l|l}
\toprule
          Metric  & ALG &     200 &      400 &      600 &      800 &     1000  \\
\midrule
     \multirow{4}{2.5cm}{$TP$} & RHCR-ECBS & 5.7±0.1 &  9.1±0.1 &  9.1±0.2 &  5.2±0.3 &  3.5±0.2  \\
        & RHCR-PBS & {\bf 6.0±0.1} & {\bf 11.1±0.1} &  6.5±0.6 &  5.2±0.5 &  3.5±0.3   \\
        & GP-R100-Re10-F2 & 5.0±0.1 &  9.1±0.1 & {\bf 11.4±0.3} & {\bf 11.5±0.6} & {\bf 10.2±0.7}   \\
\midrule
      \multirow{4}{2.5cm}{R-Time per Timestep} & RHCR-ECBS &
        \textbf{0.005±0.0} & 0.038±0.002 &  0.78±0.121 & 1.987±0.009 & 2.0±0.0   \\
        & RHCR-PBS & 0.023±0.0 &  0.29±0.016 & 1.997±0.015 &  2.01±0.006 & 2.006±0.003 \\
        & GP-R100-Re10-F2 & 0.008±0.0 & {\bf 0.011±0.001} & {\bf 0.012±0.001} & {\bf 0.012±0.001} & {\bf 0.013±0.001} \\
\midrule
      \multirow{4}{2.5cm}{R-Time per Planner Run} & RHCR-ECBS &   0.026±0.0 & 0.191±0.012 & 3.905±0.619 &  9.936±0.048 &  10.001±0.0 \\
       & RHCR-PBS & 0.114±0.002 & 1.449±0.078 & 9.985±0.074 & 10.051±0.028 & 10.03±0.013 \\
& GP-R100-Re10-F2 &   {\bf 0.008±0.0} & {\bf 0.011±0.001} & {\bf 0.012±0.001} &  {\bf 0.012±0.001} & {\bf 0.013±0.001} \\
\bottomrule
\end{tabular}
\caption{RHCR vs. Guided PIBT on Sortation centre map, with number of agents varies from 200 to 1000. It compares the Throughput ($TP$), Response Time (R-Time) per timestep, and Response Time (R-Time) per planner run.}
\label{tab:rhcr}

\end{table*}

{\bf Comparisons vs. RHCR}

\updated{The \emph{Rolling-Horizon Collision Resolution} (RHCR)~\cite{li2021lifelong} solves lifelong MAPF by decomposing
the problem into a sequence of Windowed MAPF instances,
where a Windowed MAPF solver resolves collisions among
the paths of the agents only within a bounded time horizon
and ignores collisions beyond it. In this experiment, we use implementation source codes from the original authors.\footnote{\url{https://github.com/Jiaoyang-Li/RHCR}}}

\updated{Although RHCR is a leading planner for lifelong MAPF, it has difficulty scaling to problems much larger than 1000 agents. Most of the benchmark instances in this work are larger than this size.
We thus evaluated RHCR only on our small sortation centre benchmark set, which covers the range from 200 to 1000 agents.
Table~\ref{tab:rhcr} show the throughput, response time per timestep and response time per planner run comparisons.
Note that RHCR-PBS uses PBS~\cite{ma2019searching} as MAPF planner, and RHCR-ECBS uses ECBS~\cite{barer2014suboptimal} with a suboptimality weight of 1.5.
The planning window is set to 10 and the execution window is 5, which indicates the simulator calls the MAPF planner every 5 timesteps with the bounded time horizon set to 10.}

\updated{RHCR does not call the MAPF planner every timestep. Thus we compute results for response time per timestep as total planning time divided by timesteps. Similarly, response time per planner run is the total planning time divided by the amount of MAPF planner runs. The time limit for the planner is set to 10 seconds. The results show that our Guided PIBT easily scales to a larger team size with higher throughput, and uses dramatically less computing resources.}

\subsection{Variants for Lifelong MAPF}
\updated{
In this section, we explore alternative approaches to compute the guide path using different congestion cost formulations:
\begin{itemize}
    \item \texttt{GP$_v$-R100} and \texttt{GP$_v$-R100-F2} ignores \emph{contraflow congestion} and compute guide paths on \emph{vertex congestion} only. This variant examines if \emph{vertex congestion} itself is enough for avoiding congestion.
    \item \texttt{GP$_{svc}$} using the sum of congestion costs $1 + c_e + p_v$ as objective for guide path planning. It examines if this formulation better balances congestion avoidance detour and path length.
    \item \texttt{GP$_{sui}$}. Existing study \cite{han2022optimizing} suggests adding SU-I cost-to-come congestion cost on the action cost, where the sum of added congestion cost along a path is guaranteed smaller than 1, to compute optimal length paths that tie-breaks towards less congestions. We examine this method for guiding PIBT. 
    \item \texttt{GP$_{sui}$-F2} the original SU-I cost-to-come and cost-to-go heuristic only acts as a tie-breaker, and is limited to the optimal path. Here we search for the SU-I cost-to-come minimised path within the bound of $C* \times 2$.
\end{itemize}
}

\updated{
Table \ref{tab: heiristic} compares \texttt{GP-R100} and \texttt{GP-R100-F2} with the above variants. Compared with \texttt{GP$_v$-R100} and \texttt{GP$_v$-R100-F2},  the two-part approaches (\texttt{GP-R100}) have higher throughput on most maps, showing the necessity of considering \emph{contraflow}  costs.
}

\updated{The results of \texttt{GP$_{svc}$} shows the sum-of-congestion-costs model balances path length and traffic congestion avoidance better, as it gives outstanding performance on most maps, except for the warehouse map, indicating compared with the two-part approach it underestimated the congestion in a warehouse scenario.
}

\updated{
\texttt{GP$_{sui}$} shows following an optimal length path that tie-breaking towards less congestion is not enough. By allowing suboptimal paths with costs up to $C* \times 2$, \texttt{GP$_{sui}$-F2} computes more effective guide paths.
}

\updated{Additionally, we explore alternative approaches, {\bf Traffic Cost Heuristics}, to guide PIBT agents after guide paths are computed and traffic flow are recorded on edges.}
\texttt{TH$_v$} and \texttt{TH$_v$-R100} use the recorded flow on each edge and compute vertex congestion cost minimised heuristics, where the cost on each vertex is $1 + p_v$, using Reverse Resumable A* \cite{silver2005cooperative}.
\texttt{TH$_v$} recompute all heuristics for all agents at each timestep if any agent has new tasks, as new tasks led to the change of cost-minimised single agent path and thus \updated{the cost to reach each vertex changed}.
\texttt{TH$_v$-R100} relaxed the recomputation for up to 100 heuristics per timestep, except for those with new tasks.

Table \ref{tab: heiristic} shows \texttt{TH$_v$} suffers from timeout failure on large maps, \texttt{TH$_v$-R100} suffers less but still fails on Warehouse and with lower throughput on Game. On small maps, \texttt{TH$_v$} has a slightly higher throughput than \texttt{GP$_v$-R100}, but is 5 to 10 times slower to return actions.

\begin{table*}[t]
\centering
\setlength\tabcolsep{4pt} 

\begin{tabular}{l|l|l|l|l|l|l|l|l}
\toprule
        \multirow{2}{*}{ALG} & \multicolumn{2}{c|}{Warehouse (8000)}
        & \multicolumn{2}{c|}{Sortation (600)}
        & \multicolumn{2}{c|}{Game (8000)}
        & \multicolumn{2}{c}{Room (1000)} \\
        & $TP$ & R-Time (s) &  $TP$ & R-Time (s) & $TP$ & R-Time (s) & $TP$ & R-Time (s) \\
\midrule

GP-R100 &   {\bf 23.6±5.8} &        0.18±0.083 &   10.9±3.5 &       0.004±0.002 & 45.7±8.6 &       0.153±0.041 & 2.2±1.5 &       0.003±0.006 \\

GP-R100-F2 &   22.3±5.6 &        0.168±0.08 &  10.2±3.4 &       0.002±0.002 & 45.5±8.6 &       0.138±0.036 & 3.1±1.8 &       0.003±0.004 \\

GP$_v$-R100 &   14.9±4.2 &       0.084±0.059 &  9.1±3.2 &       0.002±0.001 &   41.0±7.7 &       0.092±0.029 & 2.9±1.7 &       0.002±0.004 \\

GP$_v$-R100-F2 &   14.8±4.2 &       0.114±0.076 &  9.0±3.2 &       0.002±0.002 &  41.1±7.7 &       0.086±0.026 &   2.9±1.7 &       0.002±0.004 \\

GP$_{svc}$-R100 &   22.9±5.7 &  0.17±0.083 & {\bf 11.8±3.8} &       0.003±0.002 & {\bf 46.5±8.7} &       0.138±0.037 & 3.1±1.8 &       0.003±0.005\\

GP$_{sui}$-R100 &   20.0±5.2 &       0.076±0.029 & 8.8±3.1 &       0.002±0.001 & 44.3±8.4 &        0.06±0.015 & 2.7±1.6 &       0.002±0.002 \\

GP$_{sui}$-R100-F2 &   21.9±5.5 &       0.149±0.069 & 11.2±3.6 &       0.003±0.002 & 45.6±8.5 &       0.124±0.032 & 3.1±1.8 &       0.003±0.005\\

TH$_v$ &          timeout &                 timeout & 10.9±3.5 &        0.05±0.008 & timeout &           timeout &  {\bf 3.5±1.9} &       0.187±0.026 \\

TH$_v$-R100 &          timeout &                 timeout & 10.7±3.5 &       0.011±0.004 &  31.1±11.6 &        0.27±0.221 & 3.4±1.9 &       0.023±0.015 \\

PIBT &   19.3±5.0 &       {\bf 0.021±0.092} &  6.2±2.6 &         {\bf 0.001±0.0} & 41.9±7.7 &       {\bf 0.015±0.065} & 2.8±1.7 &       {\bf 0.001±0.002} \\
\bottomrule

\end{tabular}
\caption{Throughput ($TP$) and response time (R-Time) of vertex traffic cost heuristics (TH$_v$), vertex guide path heuristics (GPV), $1+c_e+p_{v2}$ guide path heuristics (GP$_{svc}$), SU-I cost-to-come (SUI) guide path heuristics (GP$_{sui}$), bounded SUI minimised guide path heuristics (GP$_{sui}$-F2), and two-part guide path heuristics (GP) per map, with the peak agents derived from figure~\ref{fig:allmap}. Warehouse (8000) indicates 8000 agents on the Warehouse map, the same for other maps. }
\label{tab: heiristic}
\end{table*}


\subsection{Results for One-Shot MAPF}

\begin{table}[t!]
    \centering
    \vspace*{-\baselineskip}
\begin{tabular}{c|r|rr|rrr}
\toprule
\multirow{2}{*}{Map}  & \multirow{2}{*}{Agents} &  \multicolumn{2}{c|}{RC}    &  \multicolumn{3}{c}{Solved}  \\
\cline{3-4} \cline{5-7}  & & F1.2 & F2  & F1.2 & F2 & LC* \\
\midrule
\multirow{6}{*}{\rot{Warehouse}} & 2000       &    0.982 &  1.467 &           24 &         24 &             24   \\
& 4000       &    0.971 &  1.166 &           24 &         24 &             24  \\
& 6000       &    0.942 &  1.106 &           24 &         24 &             24  \\
& 8000       &    0.852 &  1.009 &           24 &         24 &             24 \\
& 10000      &    0.770 &  0.954 &           24 &         24 &             24  \\
& 12000      &      - &    - &            0 &          0 &             24  \\
\midrule
\multirow{7}{*}{\rot{Sortation}} & 200        &    0.965 &  0.992 &           22 &         24 &             24 \\
& 400        &    0.860 &  0.888 &           22 &         24 &             24  \\
& 600        &    0.754 &  0.751 &           22 &         24 &             24  \\
& 800        &    0.728 &  0.695 &           21 &         24 &             24  \\
& 1000       &    0.759 &  0.735 &           22 &         24 &             24  \\
& 1200       &    0.815 &  0.817 &           22 &         24 &             24  \\
& 1400       &    0.903 &  0.906 &           22 &         24 &             24 \\
\midrule
\multirow{5}{*}{\rot{Game}} & 2000       &    1.058 &  1.084 &           19 &         24 &             24 \\
& 4000       &    1.070 &  1.121 &           19 &         24 &             24 \\
& 6000       &    1.072 &  1.099 &           19 &         24 &             24 \\
& 8000       &    1.029 &  1.053 &            6 &          3 &             22  \\
& 10000      &      - &    - &            0 &          0 &             14  \\
\midrule
\multirow{5}{*}{\rot{Room}} & 500        &    0.904 &  0.888 &           22 &         24 &             24  \\
& 1000       &    0.890 &  0.804 &           22 &         24 &             24  \\
& 1500       &    0.921 &  0.801 &           22 &         24 &             24 \\
& 2000       &    0.948 &  0.828 &           20 &         15 &             19  \\
& 2500       &      - &    - &            0 &          0 &              1 \\

\bottomrule
\end{tabular}
\caption{The Relative SIC (RC) of Guided LaCAM* (F1.2 and F2) to LaCAM*(LC*) on co-sovable instances. F1.2 and F2 are with guide path computed by focal search on suboptimality bound 1.2 and 2. The solved column shows the problems are solved within the 60s runtime limit. Guided LaCAM* uses 30s to compute and optimise guide paths and 30s to find a feasible solution.}
    \label{tab:lacam}
\end{table}



For one-shot MAPF, 60 seconds runtime limit is given to compute and improve solution qualities.
LaCAM* (LC*) is modified to optimise SIC.
Guided LaCAM* requires a 30 seconds setup, for computing and refining guide paths, and uses the rest of the time to 
search for a MAPF solution. If 30 seconds is not enough to compute the guide paths for all agents, we take what we have and let agents without guide paths follow their individual shortest distances.
We evaluated two versions of Guided LaCAM*, one uses \emph{Focal Search} with $w = 1.2$ to compute guide paths, labelled as \texttt{F1.2}, and the other one uses $w=2$, labelled as \texttt{F2}.  

Table \ref{tab:lacam} shows the average of the relative cost (RC), which is the $SIC_{guided}/SIC_{lacam}$. 
Our method finds solutions with higher quality in the Warehouse, Sortation and Room. 
On Warehouse, with the environment getting dense, Guided LaCAM* finds better solutions. While Sortation shows this trend ends with 90\% traversable cells occupied.
The drawback is that Guided LaCAM* only has 30 seconds left to find feasible solutions, which sacrifices the scalability if having limited runtime. 
But it concludes that for finding higher-quality solutions, it's reasonable to spend time on the computing guide path rather than relying on the anytime improvement from LaCAM*.

\section{Conclusions} \label{sec:conc}

In this work, we investigate how congestion-avoiding guide paths can improve performance for one-shot and 
lifelong MAPF.  Drawing inspiration from the literature on Traffic Assignment Problems, we develop guided variants of PIBT and LaCAM*, two recent and highly scalable MAPF planners, which nevertheless rely on ``myopic'' planning strategies.
By considering congestion costs we achieve substantial improvements in throughput for lifelong MAPF vs PIBT, 
successfully planning operations for 10,000+ agents and always returning actions for all agents in under 1 second. 
For one-shot MAPF we substantially improve solution quality for LaCAM*, at the cost of small reductions in scalability.
Future work will focus on the design of more informed and accurate objective functions for computing guide paths, as the current method relies on \emph{Focal Search} to balance between detour distance and congestion avoidance.

\newpage

\section*{Acknowledgments}
The research at Monash University was partially supported by the Australian Research Council under grants DP190100013 and DP200100025 as well as a gift from Amazon. 
The research at Carnegie Mellon University was supported by the National Science Foundation (NSF) under grant number 2328671.

We extend our special thanks to Federico Pecora from Amazon Robotics for his valuable insights and discussions that greatly contributed to this work.

\bibliography{refs}



\end{document}